    \pdfoutput=1
    \documentclass[11pt]{article}
    \usepackage{subcaption}
    \usepackage{pgfplots}
    \pgfplotsset{compat=1.18}
    
   \usepackage[final]{acl}
    
    \usepackage{times}
    \usepackage{latexsym}
    \usepackage{graphicx}
    \usepackage{booktabs}
    \usepackage{pgfplots}
    \usepackage{xcolor}
    \usepackage{float}
    
    \usepackage[T1]{fontenc}
    \usepackage[utf8]{inputenc}

    \usepackage{microtype}
    
    \usepackage{inconsolata}
    
    \usepackage{graphicx}
    
    \title{Analyzing Biases in Political Dialogue: Tagging U.S. Presidential Debates with an Extended DAMSL Framework}
    
    \author{
     \textbf{Lavanya Prahallad and Radhika Mamidi}\\
    International Institute of Information Technology, Hyderabad, India\\
    \small{
    \textbf{Email:}{lavanya.prahallad@research.iiit.ac.in and radhika@iiit.ac.in}
    }
    }
    
\begin{document}
    \maketitle
    \begin{abstract}
    We present a critical discourse analysis of the 2024 U.S. presidential debates, examining Donald Trump’s rhetorical strategies in his interactions with Joe Biden and Kamala Harris. We introduce a novel annotation framework, BEADS (Bias-Enriched Annotation for Dialogue Structure), which systematically extends the DAMSL framework to capture bias-driven and adversarial discourse features in political communication. BEADS includes a domain- and language-agnostic set of tags that model ideological framing, emotional appeals, and confrontational tactics. Our methodology compares detailed human annotation with zero-shot ChatGPT-assisted tagging on verified transcripts from the Trump–Biden (19,219 words) and Trump–Harris (18,123 words) debates. Our analysis shows that Trump consistently dominated in  key categories: Challenge and Adversarial Exchanges, Selective Emphasis, Appeal to Fear, Political Bias, and Perceived Dismissiveness. These findings underscore his use of emotionally charged and adversarial rhetoric to control the narrative and influence audience perception. In this work, we establish BEADS as a scalable and reproducible framework for critical discourse analysis across languages, domains, and political contexts.
    
    \end{abstract}
    
    \section{Introduction}
    Political Discourse Analysis (PDA) is a branch of discourse analysis that systematically investigates how language is used in political contexts to shape public opinion, gain power, gather support, and promote ideological beliefs. 
    %Through various communicative modes such as speeches, debates, interviews, and social media, politicians construct persuasive narratives not only to communicate policies but also to influence public perception, align audiences with their agendas, and undermine opposition.
    PDA uses linguistic analysis to examine both what is said and how it is said, while also taking into account the broader social and political context and the intended audience. \cite{vanDijk1997discourse, Fairclough2003}.
    
    Linguistic analysis of political language operates at multiple levels, including syntax, semantics, pragmatics, and discourse level structures. \cite{Lakoff2004, zhao2022framing} It also considers rhetorical strategies such as metaphor, repetition, presupposition, and framing, which enhance persuasive power and emotional appeal \cite{Lakoff2004}. A key concern within this analytical framework is bias, which often manifests through word choice, topic selection, speaker positioning, and emotional tone \cite{waseem2016hateful, lazaridou2020cda}. These biases can subtly promote certain viewpoints, reinforce stereotypes, and silence opposing voices ultimately shaping how people think and making fair political discussion more difficult.
    
    To systematically detect and analyze bias in political discourse, researchers employ annotation and tagging techniques. One widely used approach is the DAMSL (Dialog Act Markup in Several Layers) framework \cite{Core1997} \cite{aldayel2021unsupervised, hussain2022nlp}.

    \subsection{DAMSL framework}
    
    The DAMSL tag set was originally developed to annotate the structure and function of utterances in natural language conversations. Although initially designed for task based dialogues such as meetings or phone conversations, DAMSL has since been adapted for political discourse analysis to better capture the strategic intent, rhetorical tone, and communicative function of statements in political debates and speeches \cite{Bunt2012}. Recent studies have expanded DAMSL’s applicability to contemporary political and social domains using automated pipelines and hybrid annotation frameworks \cite{petukhova2025automated, dialogbank2018interoperable}.
    
    The original DAMSL framework included over 220 dialog act labels. However, due to its complexity, implementations such as the Switchboard DAMSL corpus reduced this to a more manageable set of 42 core tags \cite{Jurafsky1997switchboard, Core1997damsl}. These include functions such as \textit{Statement}, \textit{Question}, and \textit{Agreement}. Although these computational frameworks are effective in general dialog analysis, they often fail to capture interactive dynamics such as interruptions, challenges, and adversarial exchanges, that are central to political discourse. \cite{walker2012stance, zhang2016oxford}
    
    To address this, researchers have extended DAMSL with additional tags to reflect the characteristics of political communication. These developments have been complemented by recent annotation schemes like Dependency Dialogue Acts \cite{dda2023annotation}, which demonstrate the evolving structure of dialog tagging frameworks in capturing nuanced speaker intent. Commonly used extensions include: \textit{Statement – Opinion}, \textit{Statement – Fact}, \textit{Information Request}, \textit{Challenge / Attack}, \textit{Agreement / Disagreement}, \textit{Appeal to Emotion}, \textit{Appeal to Patriotism}, \textit{Call to Action}, \textit{Rhetorical Question}, \textit{Evade / Redirect}, \textit{Negative Personal Attack}, \textit{Justification}, and \textit{Defend / Support Claim} \cite{walker2012stance, Zhang2016conversational} \cite{yu2021framing, yu2021sotu, hu2023china}. 
    These tags have been employed in the analysis of U.S. presidential debates \cite{walker2012stance}, parliamentary sessions \cite{Ilie2010parliamentary}, and online political forums \cite{Zhang2016conversational}, enabling a more granular understanding of political argumentation and power dynamics. Tags are typically applied at the utterance or sentence level, allowing researchers to track discourse functions, conversational flow, and speaker intent across large datasets \cite{Jurafsky1997switchboard, Core1997damsl}.
    
    While the DAMSL framework and its extended adaptations provide a strong foundation for analyzing political discourse, capturing broad categories such as disagreements, challenges, and negative responses, they do not offer explicit or fine-grained tags for identifying rhetorical bias. Categories like ideological bias, rebuttal bias, dismissiveness, and adversarial verbal strategies are either only implicitly captured or entirely unrepresented. To enhance these extended tag sets, we propose a specialized set of bias-specific and adversarial tags, developed through a detailed analysis of rhetorical patterns in political debates. 
    %We applied these new tags to the 2024 U.S. presidential election debates, enabling a more precise annotation of biased rhetoric and confrontational language.

    \section{Expanding DAMSL with Bias Specific Tags}
    In this section, we introduce BEADS (Bias-Enriched Annotation for Dialogue Structure) a fine-grained annotation framework that extends DAMSL to systematically model bias, rhetorical framing, and adversarial dynamics in political dialogue. 
    %BEADS provides a structured methodology for identifying discursive strategies such as Appeals to Fear, Political Bias, and Perceived Dismissiveness, among others. 
    
    We developed BEADS tag set comprising 15 labels specifically designed to identify and quantify discourse level biases in political debate transcripts. 
    Tags address ideological framing and judgment, such as Political Bias (PB), Cognitive Bias (CB), and Cultural Bias (CBias), which capture implicit assumptions, flawed reasoning, and socially prejudiced narratives. Emotional persuasion is represented through tags like Appeals to Fear (AF), Appeals to Pride (AP), and Appeals to Patriotism (APAT), which are commonly used to influence audiences through affect rather than argument. Tags such as Gender Bias (GB) and Selective Emphasis (SE) help surface identity based and framing biases that subtly shape perception. Additionally, interactive dynamics are captured through Rebuttal (REB), Adversarial Exchange (AEX), Personal Attack (PER), Interruption (INT), and Challenge (CH), enabling the annotation of confrontation and verbal dominance. Clarification and turn-taking are also supported through tags like Correction (CORR), Seeking Explanation (SEEP), Explanation (EXPL), and Turn Request (T REQ). Importantly, this extended tag set is scalable across languages and political contexts, as it captures universal pragmatic functions and rhetorical strategies found in political speech worldwide. 
    Table~\ref{tab:bias_tags} provides illustrative definitions, generic examples, and actual excerpts from the 2024 US Presidential debates, demonstrating how each tag functions in real-world political discourse.
    %Its adaptability supports cross-cultural discourse analysis, making it valuable for global research in political communication and multilingual NLP.

    \section{Data Collection and Annotation}
    In this work, we collected official transcripts from the 2024 U.S. presidential debates: Trump–Biden (June 27,2024:CNN) and Trump–Harris (September 10, 2024: ABC News) \cite{cnn2024debate, abc2024debate}. Both transcripts were sourced from official media outlets and manually verified against the corresponding video recordings to ensure accurate speaker attribution and alignment with audience reactions. The Trump–Biden transcript contains 19,219 words and 1,450 sentences, while the Trump–Harris transcript includes 18,123 words and 1,472 sentences.
    
    To prepare the data for discourse annotation, we began by removing non-essential content such as commercial break notices, timestamps, and broadcast headers. Next, we segmented the transcripts into discrete speech units: short, self-contained segments that each convey a complete and meaningful discourse act. These units were then annotated using an extended version of the DAMSL tag set, which includes both standard dialogue act categories and additional tags specifically designed to capture features of political discourse.
    
    We ensured accurate classification of debate utterances by employing both manual and AI-assisted labeling. Discourse tags were assigned based on contextual relevance, taking into account surrounding dialogue, turn-taking patterns, and broader rhetorical strategies. This context-aware approach is essential for capturing nuanced interactions, such as extended disagreement, strategic rebuttals, or adversarial exchanges, that may be overlooked in isolated, sentence-level tagging. For instance, a sentence that appears neutral on its own may function as a rebuttal or challenge when interpreted within its conversational context. To illustrate this, Table~\ref{tab:context_examples} presents examples from the Trump–Harris debate, highlighting how tag assignments changed when context was considered.
    
    \begin{table*}[t]
    \small
    \centering
    \renewcommand{\arraystretch}{1.3}
    \begin{tabular}{|p{3.5cm}|p{2.5cm}|p{3.5cm}|p{3.8cm}|}
    \toprule
    \textbf{Sentence} & \textbf{Tag (Without Context)} & \textbf{Tag (With Context and Reasoning)} & \textbf{Context (Previous / Next Sentence)} \\
    \midrule
    "Yes, but that is not entirely true." & Statement (S) & Disagree (DIS) — response to a direct accusation about healthcare policy. & 
    \textbf{Previous:} "Your healthcare plan is leaving millions uninsured."
    \textbf{Next:} "Let me explain why that claim is misleading." \\
    \hline
    
    "We implemented tariffs to protect American jobs." & Statement (S) & Answer (ANS) — follows a question about economic policy. & 
    \textbf{Previous:} "What specific steps did you take to strengthen the economy?"
    \textbf{Next:} "These tariffs created new manufacturing opportunities." \\
    \hline
    
    "Can you explain how that’s going to work?" & Open-ended Question (OQ) & Adversarial Exchange (AEX) — asked after a vague policy proposal to expose gaps. & 
    \textbf{Previous:} "We’ll fix immigration by investing more in surveillance."
    \textbf{Next:} "That sounds good, but there’s no clarity on execution." \\
    \hline
    
    "That’s not true." & Statement (S) & Challenge (CH) — directly contradicts an opponent’s claim on immigration. & 
    \textbf{Previous:} "You opened the borders and let crime run rampant."
    \textbf{Next:} "You’re making that up just to scare people." \\
    \hline
    
    "We’ve provided support for small businesses." & Statement (S) & Rebuttal (REB) — response to an accusation of neglecting small businesses. & 
    \textbf{Previous:} "Your administration abandoned local businesses during the pandemic."
    \textbf{Next:} "In fact, we issued thousands of recovery grants." \\
    \bottomrule
    \end{tabular}
    \caption{Examples of Contextual vs. Isolated Tagging}
    \label{tab:context_examples}
    \end{table*}

    To create a reliable annotation benchmark, an expert annotator was first trained on the BEADS framework and the newly introduced bias-specific tag set. After becoming familiar with the tag definitions and contextual usage guidelines, the annotator manually labeled the debate transcripts to establish ground truth.
    
    In parallel, we used a custom-trained ChatGPT-4o model using Chain-of-Thought prompting to generate automatic annotations for the same data. The model was adapted to understand the tag schema and instructed to consider contextual cues during labeling. 
    
    We then compared the AI-generated tags against the human annotations and analyzed discrepancies to evaluate and improve the model’s accuracy. The validation revealed that 70\% of AI-generated tags matched the expert annotations. The remaining 30\% of differences were mainly due to context misinterpretation, overlapping rhetorical strategies, or subtle linguistic framing. 
    %Table~\ref{tab:validation_examples} provides examples of this comparison.
   Since this work focuses on political discourse analysis using the BEADS tag set rather than on improving the accuracy of AI generated tags, we use expert generated annotations as the definitive source of labels throughout the remainder of this work.

    \section{Analysis of Trump's Debate Strategies in the 2024 Presidential Debates}
    
    This section presents an in-depth analysis of Donald Trump's rhetorical strategies during the 2024 U.S. presidential debates against Joe Biden and Kamala Harris. Drawing on our discourse annotations and bias-specific tags, we identify recurring rhetorical patterns, contrasts in strategy, and the key factors that contributed to Trump’s perceived dominance in both debates. Across both events, Trump consistently adopted an aggressive rhetorical style, marked by frequent interruptions, confrontational language, and control over turn-taking. This combative approach disrupted the flow of argumentation, often pushing Biden and Harris into defensive or reactive positions. 
    %These behaviors align closely with the DAMSL categories of Challenge and Assertion, and correspond to the BEADS tags of Adversarial categories, reflecting Trump’s broader strategy to assert control and position himself as the dominant voice in the debate.
    
    We focus on the five most frequently observed bias-driven discourse categories in our analysis: challenges \& adverse exchanges, selective emphasis, appeals to fear, political bias, and perceived dismissiveness to understand how Trump's rhetoric shaped audience perception and debate dynamics. Table ~\ref{tab:debate_metrics} shows comparison of debate metrics between Trump vs. Biden and Trump vs. Harris.
    
    \subsection{Challenges and Adversarial Exchanges}
    A defining feature of Trump’s debate performance was his frequent use of challenges and adversarial exchanges, both of which played a crucial role in asserting dominance and destabilizing his opponents. These rhetorical strategies involved directly questioning the credibility, logic, or honesty of the other candidates, often interrupting or confronting them with accusations.
    
    Trump’s challenges often took the form of short, declarative rebuttals like “That’s not true” or sarcastic remarks aimed at undermining his opponent’s credibility. For instance, during one exchange, he quipped, “Great job, Joe look where we are now,” using sarcasm to blame Biden for perceived national decline. These confrontational statements served not only to disrupt his opponents’ arguments but also to invalidate their positions and shift focus back to himself. His adversarial exchanges often escalated into multi-turn confrontations, where he repeatedly provoked or accused until the opponent responded defensively.
    \textbf{Examples:}
    \begin{itemize}
    \item Trump vs. Biden - Challenge: \textit{"That’s a lie. You know it’s a lie."}
    \item Trump vs. Harris - Challenge: \textit{"You have no idea what you’re talking about."}
    \item Trump vs. Biden - Adversarial Exchange: \textit{"You said that, Joe. Everyone heard it. You can’t deny it now."}
    \item Trump vs. Harris - Adversarial Exchange: \textit{"You keep talking but never answer the question. That’s your problem."}
    \end{itemize}
    
    In the DAMSL-BEADS framework, such speech acts are tagged as CH (Challenge) or AEX (Adversarial Exchange), depending on intensity and discourse continuity. These tactics enabled Trump to steer the tone of the debates and put both Biden and Harris on the defensive, often making them react rather than assert their own points.
    
    \subsection{Selective Emphasis}
    A core rhetorical device in Trump's strategy was the use of selective emphasis. He repeatedly focused on national crises, from immigration to energy policy, to highlight perceived democratic failures. Rather than offering balanced arguments, Trump selectively amplified emotionally charged topics to frame his opponents as negligent or incompetent. For example, in the debate with Biden, Trump declared: "We are a failing nation," while against Harris, he asserted: "They destroyed our energy sector." These statements functioned as framing devices, guiding public attention toward blame rather than solutions.
    
    \textbf{Examples:}
    \begin{itemize}
        \item Trump (vs. Biden): \textit{"We are a failing nation."}
        \item Trump (vs. Harris): \textit{"They destroyed our energy sector."}
    \end{itemize}
    
    These statements exemplify “Selective Emphasis (SE)” and “Appeals to Fear (AF),” where rhetorical framing overrides balanced discourse.
    \subsection{Appeals to Fear}
    Trump's emotional appeals were particularly grounded in fear based rhetoric. He emphasized threats such as crime, border insecurity, and economic instability to invoke a sense of crisis, positioning himself as the only capable leader to address these issues.
    \textbf{Examples:}
    \begin{itemize}
        \item Trump (vs. Biden): \textit{"Millions of criminals are crossing the border."}
        \item Trump (vs. Harris): \textit{"They are destroying our country."}
    \end{itemize}
    
    \subsection{Political bias}
    Political bias played a central role in Trump's framing strategy. Biden, as the incumbent president, bore the brunt of Trump’s accusations on inflation, national security, and economic management. Harris, while also targeted, faced slightly less intense political bias framing, focused more on perceived inexperience and ineffective policy delivery.
    
    \subsection{Perceived Dismissiveness}
    The rhetoric was often personalized; Trump referred to Biden dismissively as "this man" and to Harris as "she," downplaying their titles and undermining their authority. Statements like "She doesn’t know what she’s doing" and "This man destroyed America" exemplified Trump's use of dismissive and delegitimizing language.
    Trump’s rhetoric included numerous instances of dismissiveness, referring to his opponents without titles or using reductive pronouns (“this man,” “she”), subtly undermining their authority and credibility.
    
    \textbf{Examples:}
    \begin{itemize}
        \item Trump (vs. Harris): \textit{"She doesn’t know what she’s doing."}
        \item Trump (vs. Biden): \textit{"This man destroyed America."}
    \end{itemize}
    
    These utterances exemplify “Perceived Dismissiveness (PD)” and represent a form of rhetorical marginalization designed to reduce the opponent’s ethos.

    \begin{table*}[htbp]
    \small
        \centering
        \renewcommand{\arraystretch}{1.4}
        \begin{tabular}{|p{3.5cm}|p{2cm}|p{2cm}|p{3.8cm}|}
            \toprule
            \textbf{Category} & \textbf{Trump vs. Biden} & \textbf{Trump vs. Harris} & \textbf{Key Difference} \\
            \midrule
             Selective Emphasis (SE) & 43 (T), 35 (B) & 40 (T), 33 (H) & Trump used SE to highlight problems and criticize Democrats, while Harris used it to express patriotic and positive messages.\\
             \hline
            Challenges (CH) & 38 (T), 31 (B) & 37 (T), 28 (H) & Biden challenged more but lacked forcefulness. \\
            \hline
            Political Bias (PB) & 29 (T), 22 (B) & 22 (T), 14 (H) & Trump blamed Biden more due to his presidency. \\
            \hline
            Adversarial Exchange (AEX) & 17 (T), 9 (B) & 13 (T), 6 (H) & More hostility in Trump vs. Biden debate. \\
            \hline
            Appeals to Fear (AF) & 32 (T), 24 (B) & 34 (T), 28 (H) & Trump used more fear based rhetoric against Harris. \\
            Personal Attacks (PER) & 21 (T), 18 (B) & 12 (T), 7 (H) & Biden faced harsher personal attacks. \\
            \hline
            Perceived Dismissiveness (PD) & 14 (T), 10 (B) & 7 (T), 3 (H) & Harris faced less dismissiveness than Biden. \\
            %\hline
            \bottomrule
        \end{tabular}
        \caption{Comparison of Debate Metrics Between Trump vs. Biden and Trump vs. Harris.}
        \label{tab:debate_metrics}
    \end{table*}

     \subsection{Qualitative Analysis}

    While Trump's core rhetorical strategies remained consistent, several quantitative and qualitative differences emerged in his interaction with each opponent, as outlined in Table ~\ref{tab:debate_metrics}. These contrasts highlight Trump’s adaptability. He adopted a more aggressive tone with Biden while using relatively less confrontational rhetoric with Harris. A combination of rhetorical strategies contributed to his perceived dominance and favorable reception among segments of the electorate.
    
    \begin{itemize}
    \item {Appeals to Fear:} Trump employed more appeals to fear in the Harris debate (34 instances) than with Biden (32), possibly exploiting her executive inexperience. 
    \item {Personalized attacks:}Trump directed more personal attacks at Biden, whose long political record and role as president made him a more immediate target. Dismissiveness was also more evident in the Biden debate, with 14 dismissive references compared to 7 in the Harris debate. Harris, on the other hand, outperformed Biden in the use of patriotic appeals, presenting a more resonant vision of national identity.
    
    \item {Political Bias (PB):}  
    Trump frequently assigned blame to his opponents by emphasizing their roles in economic decline, inflation, and foreign policy weakness. His use of “Political Bias (PB)” tags was more frequent against Biden likely due to Biden’s incumbency while Harris was framed as inexperienced and ineffective.
    
    \item {Emotional Engagement and Narrative Framing:}  
    Trump’s consistent use of fear, urgency, and crisis language evoked strong emotional reactions. By portraying himself as a decisive leader in turbulent times, he appealed to voters seeking stability.  
        \begin{itemize}
            \item \textit{"They are destroying our country."}
            \item \textit{"We are a failing nation."}
        \end{itemize}
    
    \item {Biden’s Defensive and Inconsistent Performance:}  
    Biden often appeared defensive and struggled to directly counter Trump’s assertions, weakening his rhetorical presence.  
        \begin{itemize}
            \item \textit{"The idea that somehow we are this failing country… I never heard a president talk like this before."}
        \end{itemize}
    
    \item {Harris’s Measured but Low-Impact Delivery:}  
    While Harris maintained composure and focused on unity, her lack of assertiveness and limited counter-narratives reduced the perceived impact of her performance.  
        \begin{itemize}
            \item \textit{"I believe in the promise of America."}
        \end{itemize}
    
    \item {Perceived Leadership and Command of the Stage:}  
    Trump’s assertive tone, control over turn-taking, and dominant narrative framing projected authority and leadership throughout the debates.  
        \begin{itemize}
            \item \textit{"This man destroyed America."} — an assertive frame of blame.  
            \item \textit{"We are the most admired country in the world."} — Biden’s defensive response, lacking narrative control.
        \end{itemize}  
    \end{itemize}

    \section{Results and Conclusions}
    
    To summarize, Trump effectively controlled the narrative in both 2024 presidential debates through aggressive turn-taking, selective framing, and emotionally charged messaging. His rhetorical strategies, particularly fear appeals, political bias, and dismissiveness proved more impactful than the more measured and less confrontational approaches of Biden and Harris. This calculated use of adversarial and bias-driven discourse not only shaped public perception of his opponents but also reinforced Trump’s image as a strong and decisive leader.
    
    The analysis of the debates highlights key similarities in Trump’s strategy and notable differences in how he engaged with Biden versus Harris. Across both debates, Trump employed an aggressive debate style, frequently interrupting his opponents, using fear based rhetoric (AF) to highlight crises, and framing Democratic leadership as a failure (Political Bias   PB). His dismissive language, such as referring to Harris as “she” and Biden as “this man,” reinforced his dominance in the exchanges.
    
    However, differences emerged in his approach. Trump vs. Biden was more adversarial, with Biden facing harsher personal attacks and more dismissiveness. Biden challenged Trump more often but lacked assertiveness, often appearing defensive. In contrast, Trump vs. Harris featured stronger appeals to fear, but Harris maintained composure and used Appeals to National Pride (AP) more effectively.
    
    These debate dynamics played a role in shaping voter perception. Trump’s forceful and confident presence contrasted with Biden’s defensive posture and Harris’s less aggressive challenges, reinforcing his image as a strong leader. His ability to control the debate narrative and emotionally engage voters contributed to his electoral success.
    
    %This political discourse analysis reveals that Donald Trump’s aggressive attacks, emotional appeals, and rhetorical dominance enabled him to outperform both Joe Biden and Kamala Harris in the 2024 presidential debates. His ability to control the conversation, dismiss opposing arguments, and shape the narrative contributed to his perception as a stronger and more authoritative leader. While Harris remained composed and effectively employed Appeals to National Pride (AP), her lack of assertiveness reduced the impact of her responses. Biden, though more active in issuing challenges and rebuttals, often struggled with clarity and energy, making him appear less effective in contrast to Trump. The key factors behind Trump’s advantage included his fear-based rhetoric, which resonated with anxious voters, Biden’s defensive stance, and Harris’s less confrontational style, all of which allowed Trump’s critiques to dominate. His assertiveness and narrative control reinforced his image as a decisive leader, ultimately contributing to his electoral success.
    
    In this work, we introduce BEADS—Bias-Enriched Annotation for Dialogue Structure, a trademark extension of the DAMSL framework. BEADS offers a scalable and systematic approach to annotating biased and adversarial discourse, enhancing the granularity and cross-context applicability of political dialogue analysis in multilingual NLP, political science, and media research.
    
    \section{Ethical Considerations}
    In this work, we adopt a neutral and structured approach to political discourse analysis, ensuring objectivity by avoiding partisan bias and selective framing. Debate statements were contextualized to preserve meaning, and rhetorical strategies such as Appeals to Fear (AF) were examined critically yet ethically. Candidates were assessed based solely on discourse patterns and debate performance, without personal judgment. Methodological transparency was maintained through a clearly defined annotation framework to support replicability and academic rigor.
    
    \begin{table*}[ht]
    \small
    \centering
    \renewcommand{\arraystretch}{1.3}
    \begin{tabular}{|l|p{3.2cm}|p{3.2cm}|p{3.2cm}|p{3.5cm}|}
    \hline
    \textbf{Tag} & \textbf{Description} & \textbf{Generic Example} & \textbf{Trump–Biden Example} & \textbf{Trump–Harris Example} \\
    \hline
    GB & Gender bias in language & Women are too emotional for leadership roles & — &  Trump: if elected president, Kamala Harris would be perceived by foreign leaders as "like a play toy," implying she would be easily manipulated on the world stage --Fox News interview in July 2024 \\
    \hline
    PB & Political bias in discourse & That party only cares about big donors & Trump: The other party is corrupt and incompetent. & Harris: Your party doesn’t understand working Americans \\
    \hline
    CB & Cognitive bias, flawed reasoning & No one with common sense would believe that & Trump: Everyone knows Biden's plan is a disaster & Harris: Common sense tells you this policy is wrong \\
    \hline
    AP & Appeals to pride & A true patriot would never support this policy & Trump: A true American stands with me & Harris: Proud Americans know where they stand \\
    \hline
    AF & Appeals to fear & If we don't act now, we'll face a disaster & Trump: We’re headed for destruction under Biden & Harris: We can’t afford four more years of this chaos \\
    \hline
    CBias & Displays cultural or social biases & People from that country are lazy & Trump: They send criminals over here & Harris: Systemic inequality has long been ignored \\
    \hline
    SE & Selective emphasis & They only mention the successful cases of this policy & Trump: They never talk about the failures. & Harris: They only highlight the good parts of his record \\
    \hline
    EXPL & Clarifying or explaining information. & This is why the project failed & Trump: Let me explain why I did that & Harris: Here’s why we made that choice \\
    \hline
    REB & Rebuttal or counter criticism. & That’s just a soundbite & Trump: That’s just something they told her to say & Harris: You’re repeating talking points \\
    \hline
    AEX & Adversarial exchange. & She goes again. It’s a lie & Trump: She’s lying again & Harris: You keep interrupting and lying \\
    \hline
    SEEP & Seeking explanation. & Why wasn’t anything done earlier? & Trump: Why didn’t you act sooner? & Harris: Can you explain that position? \\
    \hline
    ATTR & Blame or attribution. & He left us the worst attack on our democracy & Biden: Trump left us chaos & Harris: Trump is responsible for the mess \\
    \hline
    CORR & Correction or clarification. & I just wanna clarify here & Trump: Let me correct that & Harris: Just to clarify, that’s inaccurate \\
    \hline
    INT & Interruption. & Hold on—I wasn't finished & Trump: Let me finish! & Harris: Excuse me, I wasn’t done. \\
    \hline
    T REQ & Request for turn. & Am I allowed to respond to him? & Trump: Can I respond to that? & Harris: May I answer that, please? \\
    \hline
    \end{tabular}
    \caption{Bias-specific Dialogue Act Tag Set from the BEADS framework with examples from 2024 U.S. Presidential Debates}
    \label{tab:bias_tags}
    \end{table*}

    \bibliography{custom}

\begin{thebibliography}{24}
\providecommand{\natexlab}[1]{#1}

\bibitem[{Aldayel and Magdy(2021)}]{aldayel2021unsupervised}
Abdulaziz Aldayel and Walid Magdy. 2021.
\newblock \href {https://doi.org/10.1007/s42001-021-00101-2} {Unsupervised aspect-based stance detection in political debates}.
\newblock \emph{Journal of Computational Social Science}, 4(2):401--417.

\bibitem[{Bunt et~al.(2012)Bunt, Alexandersson, Carletta, Byron, Hwang, Lee, Romary, Purver, Stede, and Traum}]{Bunt2012}
Harry Bunt, Jan Alexandersson, Jörg Carletta, Donna~K. Byron, Kai-Yuh Hwang, Kiyong Lee, Laurent Romary, Mathew~A. Purver, Manfred Stede, and David Traum. 2012.
\newblock Iso 24617-2: A semantically-based standard for dialogue annotation.
\newblock \emph{LREC}, pages 430--437.

\bibitem[{Bunt et~al.(2018)Bunt, Petukhova, and Fang}]{dialogbank2018interoperable}
Harry Bunt, Katya Petukhova, and Alex Fang. 2018.
\newblock \href {https://doi.org/10.1007/s10579-018-9436-9} {The dialogbank: Dialogues with interoperable annotations}.
\newblock \emph{Language Resources and Evaluation}, 52(3):645--674.

\bibitem[{CNN(2024)}]{cnn2024debate}
CNN. 2024.
\newblock \href {https://www.cnn.com/presidential-debate-2024} {Cnn presidential debate (june 27, 2024)}.

\bibitem[{Core and Allen(1997{\natexlab{a}})}]{Core1997}
Mark~G. Core and James~F. Allen. 1997{\natexlab{a}}.
\newblock Coding dialogs with the damsl annotation scheme.
\newblock \emph{AAAI Fall Symposium on Communicative Action in Humans and Machines}, pages 28--35.

\bibitem[{Core and Allen(1997{\natexlab{b}})}]{Core1997damsl}
Mark~G. Core and James~F. Allen. 1997{\natexlab{b}}.
\newblock Coding dialogs with the damsl annotation scheme.
\newblock Technical Report TR 98-42, University of Rochester.
\newblock Available at University of Rochester Technical Report Archive.

\bibitem[{Fairclough(2003)}]{Fairclough2003}
Norman Fairclough. 2003.
\newblock \emph{Analyzing Discourse: Textual Analysis for Social Research}.
\newblock Routledge.

\bibitem[{Hu and Cao(2023)}]{hu2023china}
Wei Hu and Qing Cao. 2023.
\newblock \href {https://doi.org/10.1075/jlp.21092.hu} {Discourse strategies and ideological stance in chinese diplomatic speeches: A cda approach}.
\newblock \emph{Journal of Language and Politics}, 22(1):1--22.

\bibitem[{Hussain and Sajjad(2022)}]{hussain2022nlp}
Sajid Hussain and Muhammad Sajjad. 2022.
\newblock \href {https://doi.org/10.1093/llc/fqac012} {Nlp-driven analysis of imran khan’s political speeches: A case study using ericksonian patterns}.
\newblock \emph{Digital Scholarship in the Humanities}, 37(3):777--795.

\bibitem[{Ilie(2010)}]{Ilie2010parliamentary}
Cornelia Ilie. 2010.
\newblock Identity work and ideological positioning in parliamentary discourse: Conflict between adversaries in uk parliamentary debates.
\newblock \emph{Journal of Pragmatics}, 42(4):885--901.

\bibitem[{Jurafsky et~al.(1997)Jurafsky, Shriberg, and Biasca}]{Jurafsky1997switchboard}
Daniel Jurafsky, Elizabeth Shriberg, and Debra Biasca. 1997.
\newblock Switchboard dialog act corpus.
\newblock In \emph{Switchboard Dialog Act Corpus}.

\bibitem[{Lakoff(2004)}]{Lakoff2004}
George Lakoff. 2004.
\newblock \emph{Don't Think of an Elephant!: Know Your Values and Frame the Debate}.
\newblock Chelsea Green Publishing.

\bibitem[{Lazaridou et~al.(2020)Lazaridou, Koufakou, and Papadopoulos}]{lazaridou2020cda}
Angeliki Lazaridou, Angeliki Koufakou, and George Papadopoulos. 2020.
\newblock \href {https://doi.org/10.1177/0957926520939685} {A critical discourse analysis of trump’s campaign rhetoric using nlp}.
\newblock \emph{Discourse \& Society}, 31(5):497--514.

\bibitem[{News(2024)}]{abc2024debate}
ABC News. 2024.
\newblock \href {https://abcnews.go.com/presidential-debate-2024} {Abc news presidential debate (september 10, 2024)}.

\bibitem[{Nobata et~al.(2023)}]{dda2023annotation}
Chikashi Nobata et~al. 2023.
\newblock \href {https://www.researchgate.net/publication/368842950_Dependency_Dialogue_Acts_--_Annotation_Scheme_and_Case_Study} {Dependency dialogue acts -- annotation scheme and case study}.
\newblock In \emph{Proceedings of the 2023 Workshop on Dialogue Structure}.

\bibitem[{Petukhova and Kochmar(2025)}]{petukhova2025automated}
Katya Petukhova and Ekaterina Kochmar. 2025.
\newblock \href {https://arxiv.org/abs/2504.08961} {A fully automated pipeline for conversational discourse annotation: Tree scheme generation and labeling with large language models}.
\newblock \emph{arXiv preprint arXiv:2504.08961}.

\bibitem[{van Dijk(1997)}]{vanDijk1997discourse}
Teun~A. van Dijk. 1997.
\newblock \emph{Discourse as Social Interaction}.
\newblock SAGE Publications.

\bibitem[{Walker et~al.(2012)Walker, Anand, Abbott, and Grant}]{walker2012stance}
Marilyn~A. Walker, Pranav Anand, Rob Abbott, and Ricky Grant. 2012.
\newblock \href {https://aclanthology.org/D12-1054/} {Stance classification using dialogic structure}.
\newblock In \emph{Proceedings of the 2012 Joint Conference on Empirical Methods in Natural Language Processing and Computational Natural Language Learning}, pages 592--602.

\bibitem[{Waseem and Hovy(2016)}]{waseem2016hateful}
Zeerak Waseem and Dirk Hovy. 2016.
\newblock \href {https://doi.org/10.18653/v1/N16-2013} {Hateful symbols or hateful people? predictive features for hate speech detection on twitter}.
\newblock In \emph{Proceedings of NAACL-HLT}, pages 88--93.

\bibitem[{Yu and Zhao(2021)}]{yu2021sotu}
Hua Yu and Xiaoyu Zhao. 2021.
\newblock \href {https://doi.org/10.1080/17405904.2021.1933103} {A critical discourse analysis of biden’s 2021 state of the union address}.
\newblock \emph{Critical Discourse Studies}, 18(4):367--384.

\bibitem[{Yu and Ungar(2021)}]{yu2021framing}
Sheng Yu and Lyle Ungar. 2021.
\newblock \href {https://doi.org/10.18653/v1/2021.codi-1.11} {Lexical and discourse-level framing in u.s. political speech}.
\newblock In \emph{Proceedings of the 2nd Workshop on Computational Approaches to Discourse}, pages 111--121.

\bibitem[{Zhang et~al.(2016)Zhang, Culotta, and Danescu-Niculescu-Mizil}]{Zhang2016conversational}
Justine Zhang, Aron Culotta, and Cristian Danescu-Niculescu-Mizil. 2016.
\newblock Conversational flow in oxford-style debates.
\newblock In \emph{Proceedings of NAACL-HLT}.

\bibitem[{Zhang and Danescu-Niculescu-Mizil(2016)}]{zhang2016oxford}
Justine Zhang and Cristian Danescu-Niculescu-Mizil. 2016.
\newblock \href {https://doi.org/10.18653/v1/N16-1161} {Conversational flow in oxford-style debates}.
\newblock In \emph{Proceedings of NAACL-HLT}, pages 1360--1370.

\bibitem[{Zhao et~al.(2022)Zhao, Resnick, and Mei}]{zhao2022framing}
Jieyu Zhao, Paul Resnick, and Qiaozhu Mei. 2022.
\newblock \href {https://doi.org/10.18653/v1/2022.findings-acl.100} {Tracking political framing with neural discourse models}.
\newblock In \emph{Findings of the Association for Computational Linguistics: ACL 2022}, pages 1254--1265.

\end{thebibliography}
    \end{document}